# Hierarchical semantic segmentation using modular convolutional neural networks

Sagi Eppel[1]

## Abstract

Image recognition tasks that involve identifying parts of an object or the contents of a vessel can be viewed as a hierarchical problem, which can be solved by initial recognition of the main object, followed by recognition of its parts or contents. To achieve such modular recognition, it is necessary to use the output of one recognition method (which identifies the general object) as the input for a second method (which identifies the parts or contents). In recent years, convolutional neural networks have emerged as the dominant method for segmentation and classification of images. This work examines a method for serially connecting convolutional neural networks for semantic segmentation of materials inside transparent vessels. It applies one fully convolutional neural net to segment the image into vessel and background, and the vessel region is used as an input for a second net which recognizes the contents of the glass vessel. Transferring the segmentation map generated by the first nets to the second net was performed using the valve filter attention method that involves using different filters on different segments of the image. This modular semantic segmentation method outperforms a single step method in which both the vessel and its contents are identified using a single net. An advantage of the modular neural net is that it allows networks to be built from existing trained modules, as well the transfer and reuse of trained net modules without the need for any retraining of the assembled net.

## 1. Introduction.

Classification, and segmentation of objects and scenes in images is the primary task of computer vision, and is essential to any task that involves a visual understanding of the world. Many aspects of the world are hierarchical, which causes many visual recognition problems to be modular by nature [1, 2]. Examples of such problems include finding the parts of an object [3-5], or recognition of the contents of a vessel [3, 6]. In such cases, it makes sense to use a modular recognition approach, in which one method is used to identify the general object, and a second method is used to find the parts or contents of the object (Figure 1). The modular approach can considerably simplify hierarchical recognition problems by allowing each method to focus on a specific part of the task. In addition, such an approach can allow the reuse of existing methods in different parts of the task. For example, there are a large number of methods that can identify and segment general objects in images [7-13]. However, in many cases, methods for identifying specific parts or the contents of an object/vessel are not available. For such problems, it makes sense to use existing methods for recognition and segmentation of the general object, and develop specific methods for the task of recognizing the contents or parts of the object. For example, when identifying materials within glass vessels, it makes sense to use an existing off the shelf method to determine the glass vessel region and develop specific methods only for the identification of the content of the vessel [6, 14-16]. In recent years, deep learning methods based on convolutional neural networks (CNN) have dominated nearly all tasks involving image recognition, outperforming competing methods by a large margin on almost all existing benchmarks [12, 17, 18]. While a large number of CNN architectures have been suggested for various kinds of tasks, little work has been done on combining existing CNNs in a modular, hierarchical manner [19]. In general, combining neural networks in a hierarchical modular manner demands the serial connection of neural nets, such that the output of one net will be used along with the image as an input for the next net (Figure 1). Many studies have been dedicated to the combination of different architectures and layers as part of a single net that can be trained and used as one piece [20-22]. However, little work has been performed on serially connecting convolutional neural networks in a modular way, such that each net can be trained and used separately [23-25].

---
[1] sagieppel@gmail.com, Vayavision

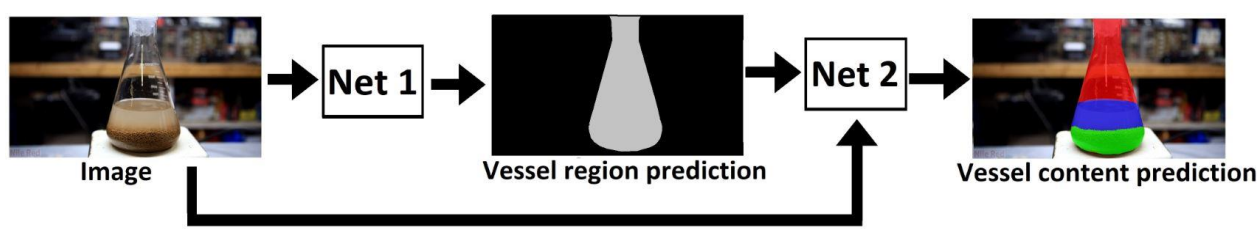

**Figure 1. Serially connected modular convolutional neural networks for hierarchical semantic segmentation. The first net recognizes and segments the vessel region. The second net uses the output of the first net and the image to identify and segment the content of the vessel. Both nets are trained separately and are combined only in the inference step.**

This work presents a modular, neural network for the task of hierarchical semantic segmentation of materials in glass vessels. One fully convolutional neural net is used to identify and segment the vessel region in the image, and a second neural net is used to identify and segment the phases of the substances within the glass vessel (Figure 1). Both of the nets are trained separately. The main challenge of the modular approach is the effective transfer of the segmentation data (vessel region) found by the first net as an input for the second net. This transfer was achieved by using the valve filter approach [26], which focuses specific features of the second net on specific segments of the image, using the segmentation map found by the first net. Comparing the results of this modular method with the results of a single step method that identifies both the vessel and its content in a single step (Figure 2), the accuracy of the hierarchical method is significantly greater than that of the single step method.

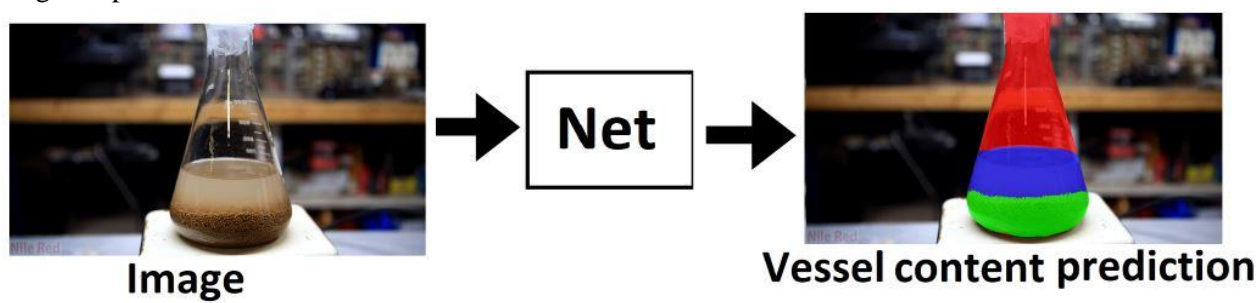

**Figure 2. Non-modular approach in which recognition of the vessel and its contents are performed using a single fully convolutional network (FCN).**

# 2. Related work

Following the success of Alexnet [18] in the Imagenet Classification Challenge in 2012, convolutional neural networks (CNNs) have emerged as the leading class of methods for image classification, detection, and segmentation [8, 17, 18, 27]. CNNs work by convolving the image with a variety of learned filters to extract a feature map, which represents the distribution of a variety of features in the image [20]. These feature maps are then used as input for the next convolution layer, and so on. Each layer extracts higher-level features, enabling better abstraction of the image contents to be achieved. Since the convolution filters used in the CNN are learned, such nets can be trained to recognize various complex categories without human aid. However, training such nets requires a large number of training images, annotated according to the specific task at hand. A number of data sets such as COCO Pascal and KADE20 are available, with thousands of pixel-level annotated images [11, 28] of general objects. In addition datasets that contain segmented parts bodies and objects are also available [3-5]. While CNNs were first demonstrated for image classification tasks [18] (hence assigning one class per image), they were soon adopted for tasks such as object detection and pixel-wise annotation [8, 12] (semantic segmentation). A large number of methods have been suggested for semantic segmentation, but most rely on some variation of the fully convolutional network (FCN) architecture [8, 12]. Focusing the attention of a CNN to a specific region of interest (ROI) of the image has also been explored in various ways ranging from using the ROI as additional input to the net to using different filters in different image regions [26]. Attention-based approaches that focus the attention of the neural net on a specific region of the image, using an attention map are another major advance of recent years.[9, 23] A similar attention map has also been created using a segmentation map that allows attention to be focused on specific segments of the image [26].

## 2.1. Modular and hierarchical neural networks

Deep neural nets are both hierarchical and highly modular. Splitting a neural network and using bottom layers (encoder) with a different set of top layers is a widely-used practice [8]. Similarly, creating a super-net that combines a large number of existing nets into a super-architecture has been applied with promising results [29, 30]. However, in all of these cases, the assembled net required retraining in order to work. In addition, replacement of one module of the net by another means that all parts of the net that come after the replaced module need to be retrained to work with the new input. Modular neural networks (MNN) solve a task by dividing it into several subtasks and training an independent neural net for each subtask [19, 21, 24, 25]. So far, this approach was mostly focused on combining prediction of several neural nets each trained as an expert for a specific subtask in a parallel and not hierarchical manner. In additional such modular neural nets were not used with convolutional neural nets for image segmentation task. CNNs are also very hierarchical by nature, wherein the first layers of the net usually extract image features such as edges and textures while higher layers identify more semantic and abstract features [20]. In addition, nets focused on more complex tasks are usually divided into different modules for different parts of the task [8, 9, 27, 31]. For example, object detections nets are usually divided into a region proposal module that identifies the general region of classification, and a refinement module that classifies the contents of this region [5, 8, 9, 27, 31-33]. However, all of these modules need to be trained together in order to work, and it is not possible to assemble several independent trained networks, or to use parts of one net in another without some retraining of the net.

# 3. Method

## 3.1. General approach

This work presents a modular hierarchical semantic segmentation approach using serially connected convolutional neural networks (CNN). The first CNN identifies the vessel region and the output of this net is used by a second CNN to identify and segment the contents of the object (Figure 1). Both segmentation steps use a standard fully convolutional network (FCN [8]), which segments the image into objects by classification of every pixel in the image into one of a given set of categories. The task examined here was the identification of the contents of a transparent vessel (Figure 1). The first step involves the segmentation of the image into the vessel and the background regions (Figure 1). A standard fully convolutional neural net, used for this task [8]. This net receives the image and outputs a vessel/background segmentation map (Figure 1). The second step of the hierarchical segmentation involves the use of a fully convolutional neural net to identify and segment the phases of the substance within the vessel (Figure 1). However, in this case, the input for the net includes both the image and the segmentation map created by the first net (Figure 1). For this purpose, it was necessary to change the net so it could use both the image and the segmentation map from the first net as an input. Addition of the segmentation map as an input to the neural net is described in section 3.3.

## 3.2. Training

The two nets, for object detection and for content detection, were trained independently and were combined only during the inference stage. The first FCN (for the recognition of the vessel) was trained using standard methods with the image as input and cross entropy between the predicted and ground truth vessel region as the loss function. The second net used for identifying the contents of the glass vessel was trained with the image and the ground truth vessel region segmentation map as an input (Figure 1). The loss was calculated as the cross-entropy between the predicted and ground truth vessel content segmentation map. Note that since the input for this net during training was the ground truth vessel region and not the predicted vessel region, it was not dependent on the first net for training. Hence, this form of training makes both nets completely independent of one another. The images and the ground truth annotations (segmentation maps) were taken from the dataset of materials in transparent vessels described in Section 4. Since both nets were trained independently, neither of the nets had the chance to adjust to the other during training. Hence, their combination during the testing/inference step is basically the combination of two independent nets. To evaluate the results of the hierarchical modular approach vs. a single step approach, a third fully convolutional neural (FCN) net was trained to predict the vessel region and its contents in a single stage (Figure 2). The training loss was found using the ground truth semantic segmentation map of the vessel content, taken from the materials in the vessels dataset (Section 4).

## 3.3. Use of the segmentation map as input

The valve filters method [26], is based on the idea that when analyzing an image segmented into different known regions it is desirable to extract different information from different regions. Convolutional neural nets extract information from images by convolving a set of filters with the image, whereby each filter extracts a different feature. Extraction of different information from different image regions can therefore be performed by applying different filters to different segments of the image. A more moderate approach is to apply the same set of filters to the entire image but to assume that different filters will have different relevance in different regions. For example, in case of an image segmented into two regions classified as object and background, some features might be more relevant to the object region in the image while others might be more relevant to the background region. The valve filters are used to inhibit specific features in a specific region of the image based on the segmentation map (Figure 3).

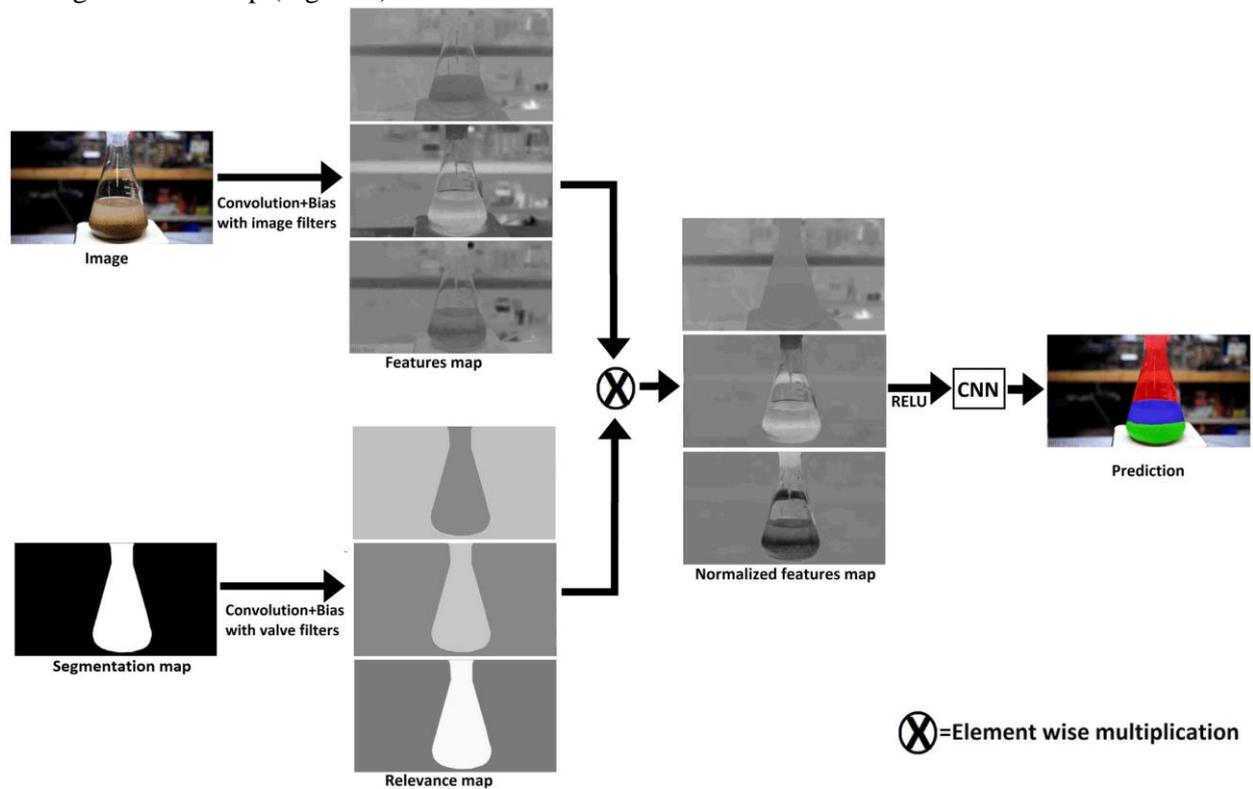

**Figure 3. The valve-filter approach for using the segmentation map as input to convolutional neural nets. The image and the segmentation map are each passed through a separate convolution layer to give a feature map and a relevance map, respectively. Each element in the feature map is multiplied by the corresponding element in the feature map to give a normalized feature map that is passed as input to the next layer of the net.**

The valve filters act as a kind of valve that regulates the activation of the corresponding image filter in the different segments of the image. Applying the valve filters was performed as shown in Figure 3, using the following steps:
a) A set of filters are convolved with the image to generate a feature map;
b) For each filter applied to the image, a corresponding valve filter exists that is convolved with the segmentation mask to give the relevance map;
c) Each element in the feature map is multiplied by the corresponding element in the relevance map to give a normalized feature map;
d) The normalized featured map is used (after RELU) as input for the next layer. Hence, the relevance map is used to inhibit or enhance the activation of the features depending on whether they appear in the ROI or in the background region.

# 4. Data set for materials in glass vessels

The task of modular semantic segmentation was examined using the dataset of materials in glass vessels. The main purpose of this dataset is to enable the training of neural networks in the task of recognizing the physical and chemical properties of substances inside glassware vessels in a laboratory setting. The handling of materials inside glassware or other transparent vessels is the main activity in most chemistry laboratory work, and is essential for a wide range of methods used in materials research [26, 34-36]. The dataset includes a thousand images of substances in different physical phases inside laboratory glassware [26]. Each image in the data set is supplied with pixel-wise annotation, according to several levels of classification (Figure 4). The first level segments the image into the glassware region and the background region; the second layer segments the vessel region into filled and empty regions; the third layer segments the vessel contents into liquid and solid phases; and the final level gives exact categories for the phase of the material in the vessel, including liquid, solid, powder, foam, vapor, suspension and emulsion. Each set of categories was considered as a different task and a separate net was trained for each task. The ground truth of the vessel/background annotation was used to train the first modular net for the vessel region detection (Figure 4). The rest of the ground truth annotations were used to train the second modular net (for detection of the vessel contents), where a different net was trained for each level. The images from this dataset were collected from YouTube channels dedicated to chemistry experiments (mainly NileRed, NurdeRage, ChemPlayer), and were manually labeled by Alexandra Emanuel and Mor Bismuth.

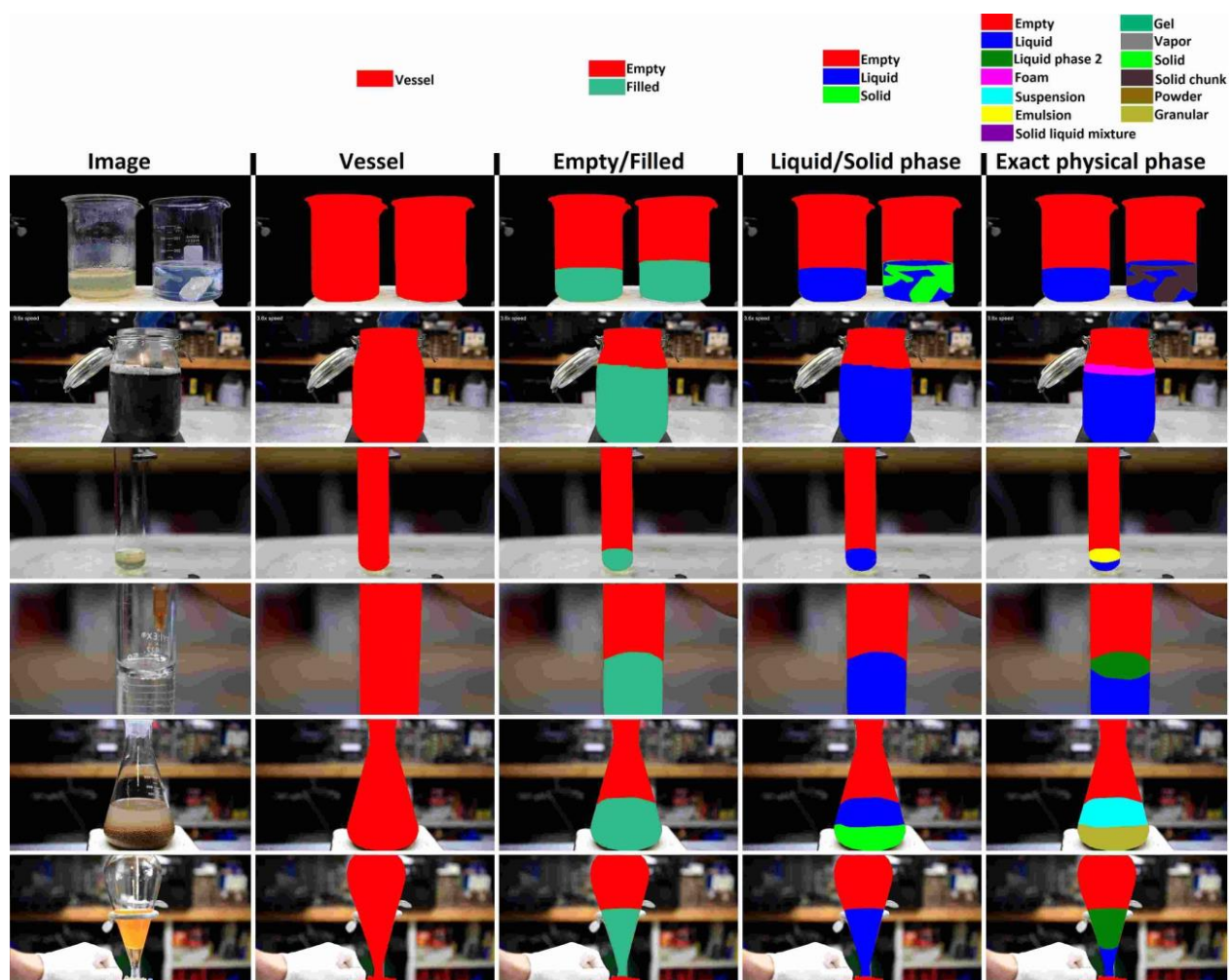

**Figure 4. Images from the data set of materials in glass vessels and their annotation. Each column in the image displays a different level of annotation.**

## 5. Results and discussion

The results of both the hierarchical method and the single step method are shown in Figure 5 and Table 1. Quantitative evaluation of the accuracy of each method was performed by calculating the intersection over union (IOU) of the prediction and ground truth for each pixel. It can be seen from the results that the hierarchical modular approach gave better accuracy compared with the single net method. This was the case for all classes and for all levels of classification. The specific architecture of the net used is not likely to affect the results since both the one net mode and the hierarchical mode use standard FCN architecture. It can be argued that connecting two neural networks in a series creates one net of twice the depth and twice the number of parameters as the original net, and that the extra depth and additional parameters are responsible for the increase in accuracy. However, if the serially connected nets were treated as a very deep single net, this net would be too deep to be trained effectively. Hence, while the combined modular nets work efficiently as one segment, they must be used and trained as independent nets. The modular net also allows the replacement of individual subnets by alternative methods which perform the same function. For example, the net trained for identifying the vessel content could be used with any method that supplies the vessel region segmentation as input. To examine this property, the vessel region prediction by the first net was replaced by the ground truth vessel region as input to the second net. This led to far more accurate results compared with the method that use of the output of the vessel detection net as an input (Table 1).

**Table 1. Intersection over union results (IOU) for vessels content recognition.**

|  | Single net[a] | Modular net two steps[b] | Modular net with ground truth vessel region[c] |
|---|---|---|---|
| **Vessel region** | | | |
| Background | 94.43% | 94.27% | 99.65% |
| Vessel | 83.92% | 83.45% | 98.91% |
| **Fill level** | | | |
| Background | 93.42% | 94.26% | 99.58% |
| Empty | 65.27% | 67.00% | 82.32% |
| Filled | 69.67% | 73.43% | 82.07% |
| **Solid/Liquid** | | | |
| Background | 93.44% | 94.28% | 99.61% |
| Empty | 61.77% | 68.23% | 83.59% |
| Liquid | 59.44% | 69.65% | 77.90% |
| Solid | 47.20% | 60.16% | 62.26% |
| **Exact physical phase** | | | |
| Background | 92.76% | 94.30% | 99.57% |
| Vessel | 59.80% | 66.08% | 80.15% |
| Liquid | 43.79% | 51.17% | 58.08% |
| Liquid phase two | 16.61% | 21.81% | 21.95% |
| Suspension | 27.16% | 28.92% | 29.18% |
| Foam | 5.69% | 6.73% | 4.72% |
| Solid | 0.22% | 8.91% | 8.90% |
| Powder | 24.56% | 27.98% | 26.54% |
| Granular | 27.71% | 25.45% | 28.78% |
| Bulk | 2.38% | 9.03% | 8.48% |

a. Single net FCN recognition, the vessel and its contents found by a single FCN.
b. Modular net, vessel region and vessel contents found by two different nets.
c. Modular net, using ground truth vessel region as input for content recognition net.

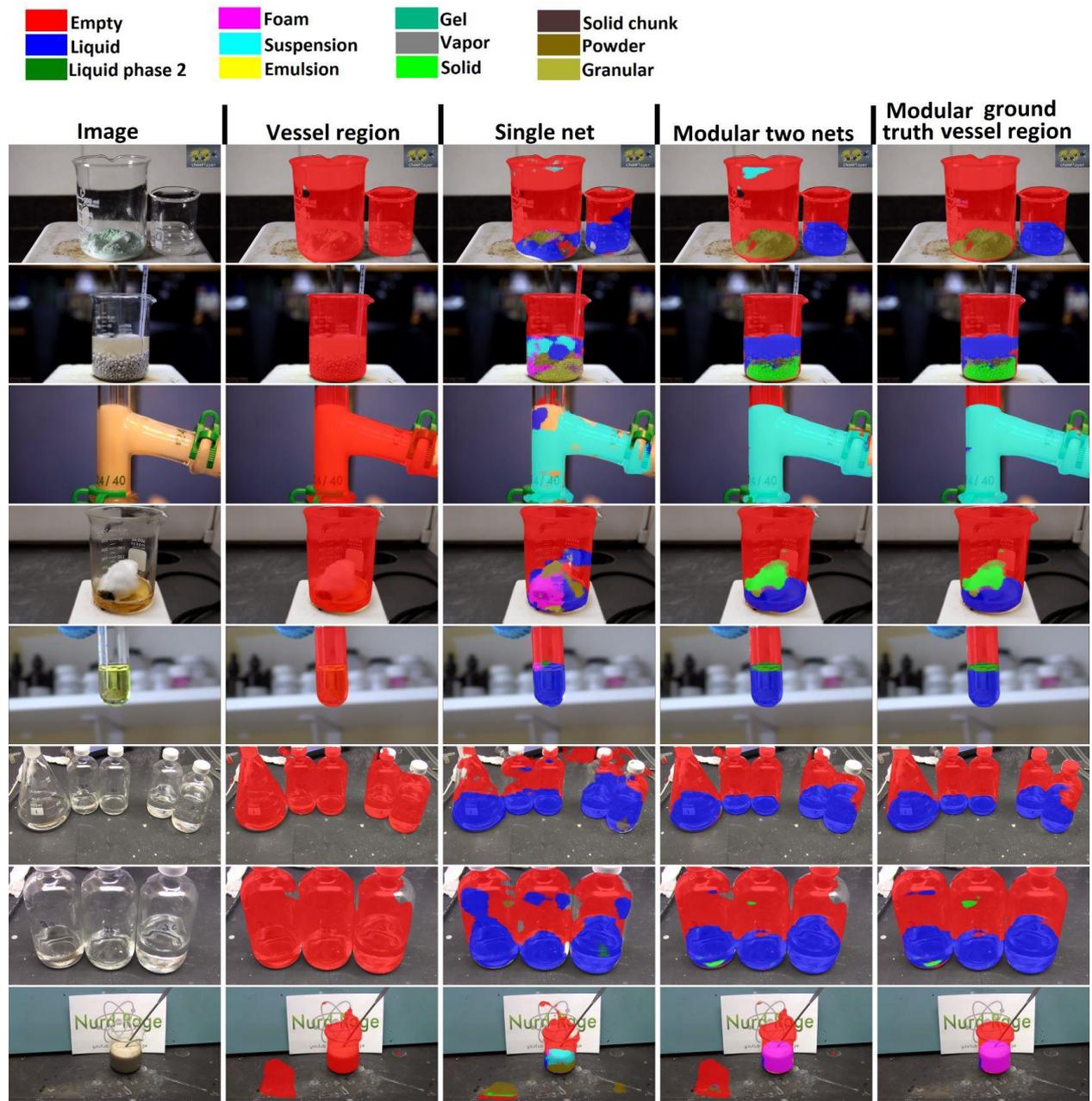

**Figure 5. Results predictions of various methods for semantic segmentation of the contents of glass vessels.**
**a. Single net FCN recognition. The vessel and its contents found by a single FCN.**
**b. Modular net, vessel region and vessel contents found by two different nets.**
**c. Modular net, using ground truth vessel region as input for vessel content recognition net.**

# 6. Conclusion

This work demonstrated that it is possible to combine two independent neural networks, trained for different but complementary tasks, in order to solve higher level hierarchical recognition problems. The problem studied was the applied hierarchical segmentation of the contents of a transparent vessel. The results show that the serial combination of two nets independently trained to perform subtasks of the main recognition task (identifying the vessel region and identifying the transparent container contents) gave a more accurate result when compared with a single net trained to perform the full recognition problem (identifying the vessel and its contents). This result suggests that a modular approach can improve accuracy when considering hierarchical recognition problems such as segmenting parts of a body or contents of a vessel. Another major benefit of the modular approach is that it allows the use of preexisting methods in each part of the hierarchical recognition, and in this way allows the use of existing nets for new tasks without the need for retraining.  Similarly, since each net in the modular approach was trained independently for a different task, each net can be reused for other tasks without the need for retraining. Finally increasing the depth of neural nets has been shown to be one of the most important factors for improving prediction accuracy. However, very deep nets are very hard to train due to the vanishing gradient problem. The modular approach combines two nets serially and creates a de facto net with double the depth of each separate net. At the same time, the fact that each net is trained separately solves the depth issue during training.

# 7. Thanks

Thanks are due to Alexandra Emanuel and Mor Bismuth for their work on the labeling of the data set, and to the creators of the Youtube channels NileRed, NurdeRage and ChemPlayer for allowing the use of frames from their videos in the creation of the data set and this paper.

# 8. Supporting material

Code for the FCN valve filter input (used for vessel content detection) can be found in:
https://github.com/sagieppel/Focusing-attention-of-Fully-convolutional-neural-networks-on-Region-of-interest-ROI-input-map-
Code for standard FCN with used for the vessel region detection net net can be downloaded from:
https://github.com/sagieppel/Fully-convolutional-neural-network-FCN-for-semantic-segmentation-Tensorflow-implementation
The dataset for materials in vessels can be found in: https://github.com/sagieppel/Materials-in-Vessels-data-set

# 9. References

1. Riesenhuber, M. and T. Poggio, *Hierarchical models of object recognition in cortex.* Nature neuroscience, 1999. **2**(11).
2. Zhen, Z., H. Fang, and J. Liu, *The hierarchical brain network for face recognition.* PloS one, 2013. **8**(3): p. e59886.
3. Chen, X., et al. *Detect what you can: Detecting and representing objects using holistic models and body parts*. in *Proceedings of the IEEE Conference on Computer Vision and Pattern Recognition*. 2014.
4. Gong, K., et al., *Look into Person: Self-supervised Structure-sensitive Learning and A New Benchmark for Human Parsing.* arXiv preprint arXiv:1703.05446, 2017.
5. Tsogkas, S., et al., *Semantic part segmentation with deep learning.* CoRR, abs/1505.02438, 2015.
6. Eppel, S., *Tracing the boundaries of materials in transparent vessels using computer vision.* arXiv preprint arXiv:1501.04691, 2015.
7. Chen, L.-C., et al., *Deeplab: Semantic image segmentation with deep convolutional nets, atrous convolution, and fully connected crfs.* arXiv preprint arXiv:1606.00915, 2016.

8. Long, J., E. Shelhamer, and T. Darrell. *Fully convolutional networks for semantic segmentation*. in *Proceedings of the IEEE Conference on Computer Vision and Pattern Recognition*. 2015.
9. Chen, L., et al., *SCA-CNN: Spatial and Channel-wise Attention in Convolutional Networks for Image Captioning.* arXiv preprint arXiv:1611.05594, 2016.
10. Badrinarayanan, V., A. Kendall, and R. Cipolla, *Segnet: A deep convolutional encoder-decoder architecture for image segmentation.* arXiv preprint arXiv:1511.00561, 2015.
11. Zhou, B., et al., *Semantic understanding of scenes through the ADE20K dataset.* arXiv preprint arXiv:1608.05442, 2016.
12. Garcia-Garcia, A., et al., *A Review on Deep Learning Techniques Applied to Semantic Segmentation.* arXiv preprint arXiv:1704.06857, 2017.
13. Caesar, H., J. Uijlings, and V. Ferrari, *COCO-Stuff: Thing and Stuff Classes in Context.* arXiv preprint arXiv:1612.03716, 2016.
14. Eppel, S. and T. Kachman, *Computer vision-based recognition of liquid surfaces and phase boundaries in transparent vessels, with emphasis on chemistry applications.* arXiv preprint arXiv:1404.7174, 2014.
15. Hambrice, K. and H. Hopper, *A Dozen Ways to Measure Fluid Level and How They Work.* Sensors-the Journal of Applied Sensing Technology, 2004. **21**(12): p. 14-19.
16. Eppel, S., *Tracing liquid level and material boundaries in transparent vessels using the graph cut computer vision approach.* arXiv preprint arXiv:1602.00177, 2016.
17. Dumoulin, V. and F. Visin, *A guide to convolution arithmetic for deep learning.* arXiv preprint arXiv:1603.07285, 2016.
18. Krizhevsky, A., I. Sutskever, and G.E. Hinton. *Imagenet classification with deep convolutional neural networks*. in *Advances in neural information processing systems*. 2012.
19. Azam, F., *Biologically inspired modular neural networks*, 2000.
20. Gonzalez-Garcia, A., D. Modolo, and V. Ferrari, *Do semantic parts emerge in Convolutional Neural Networks?* arXiv preprint arXiv:1607.03738, 2016.
21. Watanabe, C., K. Hiramatsu, and K. Kashino, *Modular Representation of Layered Neural Networks.* arXiv preprint arXiv:1703.00168, 2017.
22. Zhao, B., et al., *A survey on deep learning-based fine-grained object classification and semantic segmentation.* International Journal of Automation and Computing, 2017: p. 1-17.
23. Chen, K., et al., *ABC-CNN: An attention based convolutional neural network for visual question answering.* arXiv preprint arXiv:1511.05960, 2015.
24. Chen, K., *Deep and modular neural networks*, in *Springer Handbook of Computational Intelligence*. 2015, Springer. p. 473-494.
25. Auda, G. and M. Kamel, *Modular neural networks: a survey.* International Journal of Neural Systems, 1999. **9**(02): p. 129-151.
26. Eppel, S., *Setting an attention region for convolutional neural networks using region selective features, for recognition of materials within glass vessels.* arXiv preprint arXiv:1708.08711, 2017.
27. Ren, S., et al. *Faster R-CNN: Towards real-time object detection with region proposal networks*. in *Advances in neural information processing systems*. 2015.
28. Lin, T.-Y., et al. *Microsoft coco: Common objects in context*. in *European conference on computer vision*. 2014. Springer.
29. Shazeer, N., et al., *Outrageously large neural networks: The sparsely-gated mixture-of-experts layer.* arXiv preprint arXiv:1701.06538, 2017.
30. Fernando, C., et al., *Pathnet: Evolution channels gradient descent in super neural networks.* arXiv preprint arXiv:1701.08734, 2017.
31. Shih, K.J., S. Singh, and D. Hoiem. *Where to look: Focus regions for visual question answering*. in *Proceedings of the IEEE Conference on Computer Vision and Pattern Recognition*. 2016.
32. Kantorov, V., et al. *Contextlocnet: Context-aware deep network models for weakly supervised localization*. in *European Conference on Computer Vision*. 2016. Springer.
33. Gonzalez-Garcia, A., D. Modolo, and V. Ferrari, *Objects as context for part detection.* arXiv preprint arXiv:1703.09529, 2017.

34. Ley, S.V., et al., *Camera-enabled techniques for organic synthesis.* Beilstein journal of organic chemistry, 2013. **9**: p. 1051.
35. Ley, S.V., et al., *Organic synthesis: march of the machines.* Angewandte Chemie International Edition, 2015. **54**(11): p. 3449-3464.
36. Fitzpatrick, D.E., C. Battilocchio, and S.V. Ley, *Enabling Technologies for the Future of Chemical Synthesis.* ACS central science, 2016. **2**(3): p. 131-138.